\pdfoutput=1

\documentclass[11pt]{article}

\usepackage[preprint]{acl}

\usepackage{times}
\usepackage{latexsym}
\usepackage{amsmath}
\usepackage{listings}
\usepackage{color}
\usepackage{xcolor}
\usepackage{caption}
\usepackage[T1]{fontenc}
\usepackage[utf8]{inputenc}
\usepackage{microtype}
\usepackage{inconsolata}
\usepackage{graphicx}

\title{\textit{MALT: Mechanistic Ablation of Lossy Translation} in LLMs for a Low-Resource Language: Urdu}

\author{Taaha Saleem Bajwa \\
  Independent\\
  \texttt{taaha.s.bajwa@gmail.com} 
  }

\begin{document}

\maketitle

\begin{abstract}
LLMs are predominantly trained on English data, which leads to a significant drop in performance on low-resource languages. Understanding how LLMs handle these languages is crucial for improving their effectiveness. This study focuses on Urdu as a use case for exploring the challenges faced by LLMs in processing low-resource languages. LLMs primarily reason in English when prompted in another language, with the final layers acting as translators to convert the English response into the target language. This study finds that even for low-resource languages, the internal latent response of LLMs in English is quite coherent; however, the translation features are lossy and result in poor translations, leading to reduced performance. By mechanistically removing these translation features and using a separate translation model to translate the LLM’s internal latent response, the performance of LLMs improves significantly while also preserving the cultural nuances of the input in low-resource languages.
\end{abstract}

\begin{figure*}[t]
  \centering
  \includegraphics[width=\textwidth]{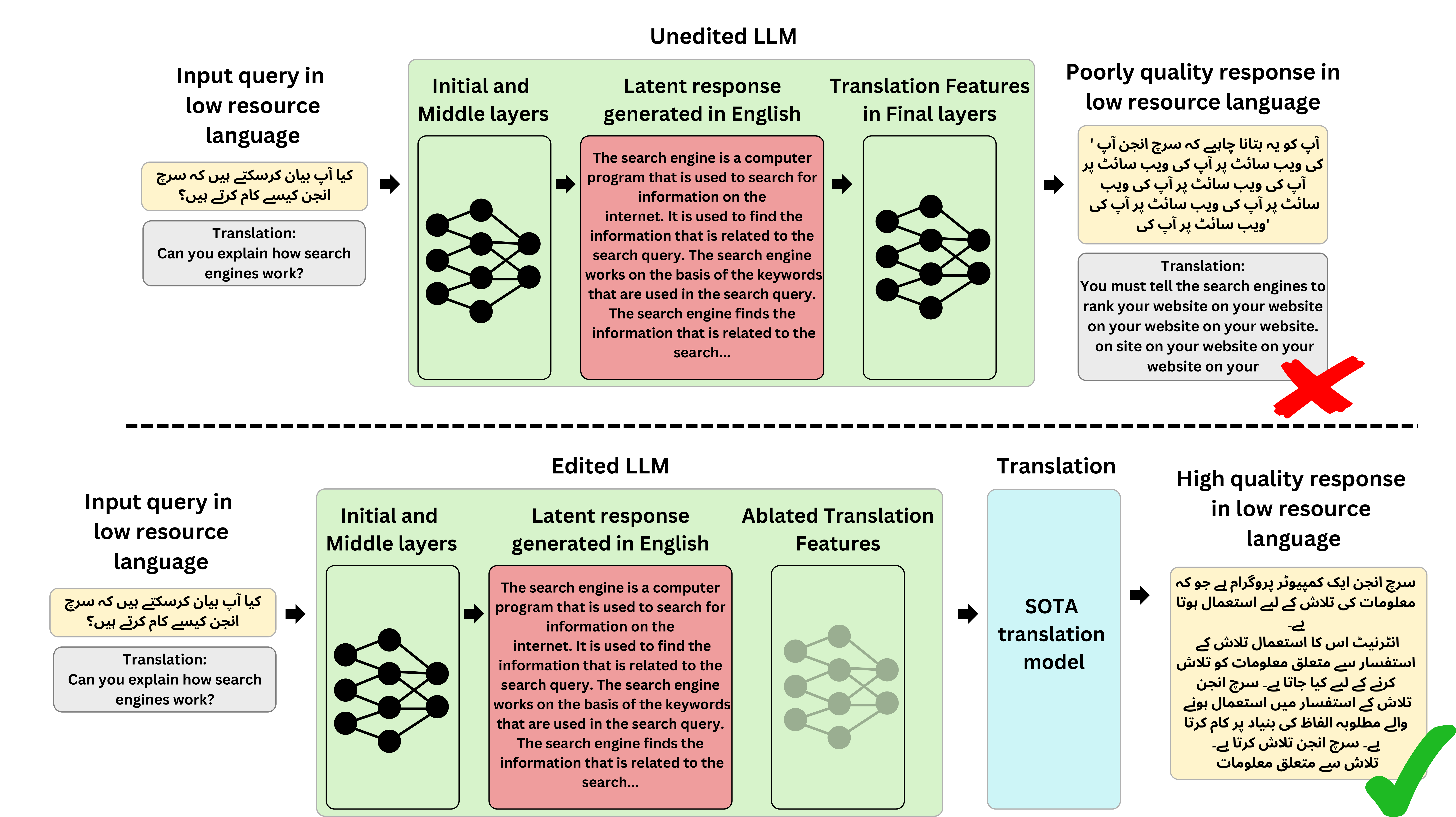}
  \caption{(Above) Baseline LLM operation for low-resource languages, showing poor translation quality due to lossy translation features in the final layers. (Below) Proposed method, where lossy translation features are removed from the final layers and replaced with a dedicated machine translation model, resulting in high-quality responses. \textit{Note: The gray boxes are included solely to aid reader comprehension and are not part of the methodology.}}
  \label{fig:experiments}
\end{figure*}

\footnote{Results along with code are available here \url{https://github.com/taaha/MALT}}

\section{Introduction}

Most Large Language Models (LLMs) are trained on English-dominant corpora. Even multilingual LLMs, like Llama 3, contain only around 5\% non-English data \citep{Llama_3}, which results in a significant performance gap across different languages. Furthermore, this multilingual capability is generally focused on high-resource languages such as French and German, leaving limited support for low-resource languages. This greatly restricts universal AI accessibility. For example, Urdu, a low-resource language, is spoken by 230 million people \citep{Hussain_Hussain_2022}, yet receives minimal representation in LLM training data. Addressing these gaps is important to provide fairer AI access for people who speak different languages around the world.

Previous studies have suggested that, due to their English-centric training datasets, LLMs use English as their latent language even when prompted in another language \citep{wendler2024llamasworkenglishlatent}. It is only in the final layers that LLMs translate the response from this latent language into the language of the input prompt. The capability of LLMs to process non-English languages relies mostly on a very small number of neurons located in the final and initial layers \citep{tang2024languagespecificneuronskeymultilingual}, which are primarily involved in translating the prompt to and from the latent language. This study aims to reinforce this hypothesis for low-resource languages and address the translation losses that are particularly significant for these languages.

Mechanistically erasing a direction within a layer has been shown to alter the performance of LLMs \citep{arditi2024refusallanguagemodelsmediated}. Additionally, studies have demonstrated that by modifying language-specific neurons, it is possible to control target language generation \citep{kojima2024multilingualabilitydecoderbasedpretrained}. Building on this, this study shows that for smaller LLMs, regardless of the task or prompt, low-resource language generation is primarily mediated by a single direction in the final layers, which can be easily removed mechanistically.

By mechanistically removing the translation features in the final layer of the LLM, internal latent responses in English are obtained that are generally more coherent than the outputs generated by the unedited LLM in the target language. These responses can then be translated into low-resource languages using state-of-the-art (SOTA) machine translation models. This highlights a key insight into how LLMs process low-resource languages: LLMs are effective at understanding these languages but struggle to generate coherent responses in them.

One traditional solution for obtaining high-quality responses from LLMs is to translate the input into English, feed the English text to the model, and then translate the generated output back into the original language. However, this approach often fails to capture the cultural nuances of the input text. The method in this study addresses this issue by keeping the input in the target language and only replacing the lossy translations in the final layer with more accurate ones. This approach preserves the cultural context of the target language.

While the output shows improved coherence and represents a step forward in performance, there is still significant room for further enhancement in handling low-resource languages, and additional efforts are needed to achieve even better results. 

The contributions of this study to the field of LLMs for low-resource languages are as follows:

\begin{enumerate} 
\item This study advances our understanding of how LLMs process low-resource languages by demonstrating that LLMs are significantly better at understanding low-resource languages than generating responses in them. 
\item A novel method is proposed to enhance the performance of LLMs on low-resource languages without requiring pre-translation of input prompts (i.e., input prompts remain in the target language). This approach also ensures that cultural nuances are preserved, as the input retains its original linguistic and contextual integrity.
\end{enumerate}

To the best of the author's knowledge, no similar prior study has been conducted on understanding or improving the performance of LLMs for low-resource languages.

Throughout this paper, the standard operation of an LLM, where the model processes input in a low-resource language without any modifications, is referred to as the Baseline. This terminology will be consistently used to distinguish between unedited LLM performance and the proposed methodology, MALT.

\section{Methodology}

\subsection{Dataset}
For this study, a diverse dataset of questions was generated using ChatGPT\cite{chatgpt}. The selected questions are primarily those that require explanatory answers spanning several lines, as the main focus of this study is to assess both the fluency and relevance of answers generated by LLMs in low-resource languages. Each question is provided in English along with a corresponding Urdu translation. The total dataset consists of 239 questions, of which 15 are used to identify translation features, while the remaining 223 are used for evaluation.

\subsection{Models}
Due to computational limitations, this study is conducted on smaller LLMs with 2 to 3 billion parameters. These LLMs require less compute power and generally perform poorly on low-resource languages, making them an ideal testbed to understand and improve the processing of such languages. For this study, we use Gemma-2-2b \citep{gemmateam2024gemma2improvingopen} and Llama-3.2-3b \citep{dubey2024llama3herdmodels}. A brief overview of these models is provided in Table \ref{table:LLMs}.

For machine translation of English outputs generated from the edited LLM into Urdu, there are numerous options available, with models ranging in size from a few hundred MBs to several GBs. For higher translation accuracy, We use a fine-tuned Urdu-specific checkpoint \citep{finetuned_mbart_model_card} of the large mBART machine translation model \citep{tang2020multilingual}.

\begin{table}[!h]
  \centering
  \begin{tabular}{lccc}
    \hline
    \textbf{Model} & \textbf{Parameters} & \textbf{Layers} \\
    \hline
    Gemma-2-2b & {2.6B} & {26} \\
    Llama-3.2-3b & {3.2B} & {28} \\
    \hline
  \end{tabular}
  \caption{LLMs used for MALT.}
  \label{table:LLMs}
\end{table}

This study utilized nearly 90 GPU hours on an RTX A4000.

\subsection{Finding Translation Direction Across Layers}

To identify translation direction, we use a method inspired by \cite{arditi2024refusallanguagemodelsmediated}. The model is prompted with \( N \) Urdu and English questions separately, while caching their residual activations for each layer. The value of \( N = 16 \) is used in this study.

Let \( \mathbf{R}_{\text{eng}, \ell} \) represent the cached residual activations for English questions and \( \mathbf{R}_{\text{urd}, \ell} \) represent the cached residual activations for Urdu questions at layer \( \ell \). This layer \( \ell \) is one of the final layers and upon hit and trial, its optimal value was found to be \( \ell \) = 24 for Gemma-2-2b and \( \ell \) = 25 for Llama-3.2-3b.

The mean residual activations for each language at layer \( \ell \) are computed as follows:

\begin{equation}
\mathbf{m}_{\text{eng}, \ell} = \frac{1}{N} \sum_{i=1}^{N} \mathbf{R}_{\text{eng}, \ell, i}
\end{equation}

\begin{equation}
\mathbf{m}_{\text{urd}, \ell} = \frac{1}{N} \sum_{i=1}^{N} \mathbf{R}_{\text{urd}, \ell, i}
\end{equation}

Next, the mean residual activations are subtracted, followed by normalization to find the translation direction for layer \( \ell \):

\begin{equation}
\mathbf{d}_{\ell, \text{norm}} = \frac{\mathbf{m}_{\text{urd}, \ell} - \mathbf{m}_{\text{eng}, \ell}}{\|\mathbf{m}_{\text{urd}, \ell} - \mathbf{m}_{\text{eng}, \ell}\|}
\end{equation}

\subsection{Removing Translation Direction}

For each residual activation \( \mathbf{R} \), the translation direction is ablated by first computing the projection onto the direction \( \mathbf{d}_{\ell, \text{norm}} \), scaling it, and then subtracting the scaled projection:

\begin{equation}
\mathbf{R}_{\text{ablation}} = \mathbf{R} - \left( (\mathbf{R} \cdot \mathbf{d}_{\ell, \text{norm}}) \cdot \mathbf{d}_{\ell, \text{norm}} \right)
\end{equation}

This procedure ablates the translation features.

\section{Experimental Results}

\subsection{Error types}
We see following different type of erroneous responses from edited LLMs.
\begin{enumerate} 
\item Fluency error: The response from the edited LLM is incoherent and unreadable, consisting of random characters. 
\item Repetition error: The response from the edited LLM consists only of multiple repetitions of the query. 
\item Non-relevant error: The response is coherent but does not relate to the query or does not answer the query effectively. \end{enumerate}
Examples of above errors can be seen in Appendix \ref{sec:appendix}.

These errors may be caused due to the fact that most neurons are polysemantic \citep{scherlis2023polysemanticitycapacityneuralnetworks} and some other features are also ablated when removing translation direction. Upon observation, it seems that some non-relevant errors are also caused due to poor understanding of the input prompt by LLMs.

\subsection{Evaluation}

\begin{figure}[t]
  \includegraphics[width=\columnwidth]{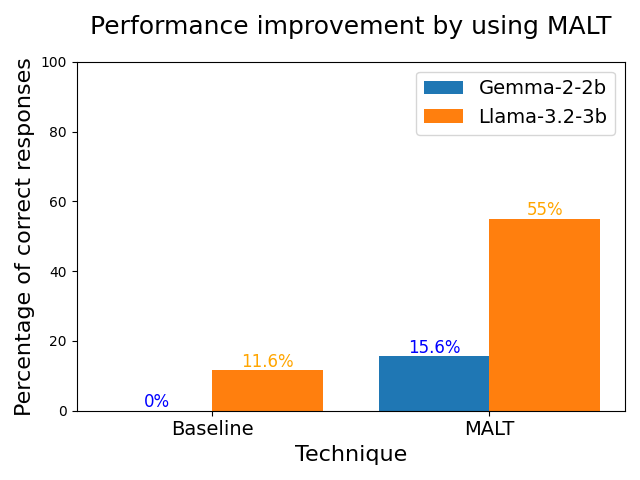}
  \caption{Percentage of correct responses for Baseline and MALT in Gemma-2-2b and Llama-3.2-3b.}
  \label{fig:experimental_result}
\end{figure}

The results are evaluated by a human fluent in Urdu and English and only those responses are considered correct which are both fluent and relevant to the input query. Evaluation are done on english latent response prior to translation in order to ignore errors due to translation as they are largely dependant on translation model used.

Figure~\ref{fig:experimental_result} shows the percentage of correct responses for two models, Gemma-2-2b and Llama-3.2-3b, under baseline and MALT conditions. Llama-3.2-3b achieved a significant improvement from 11.6\% at baseline to 55\% with MALT, while Gemma-2-2b increased from 0\% to 15.6\%, indicating that MALT effectively increases LLM performance on low-resource languages.

\subsection{Are cultural nuances preserved?}

\begin{figure}[!h]
  \includegraphics[width=\columnwidth]{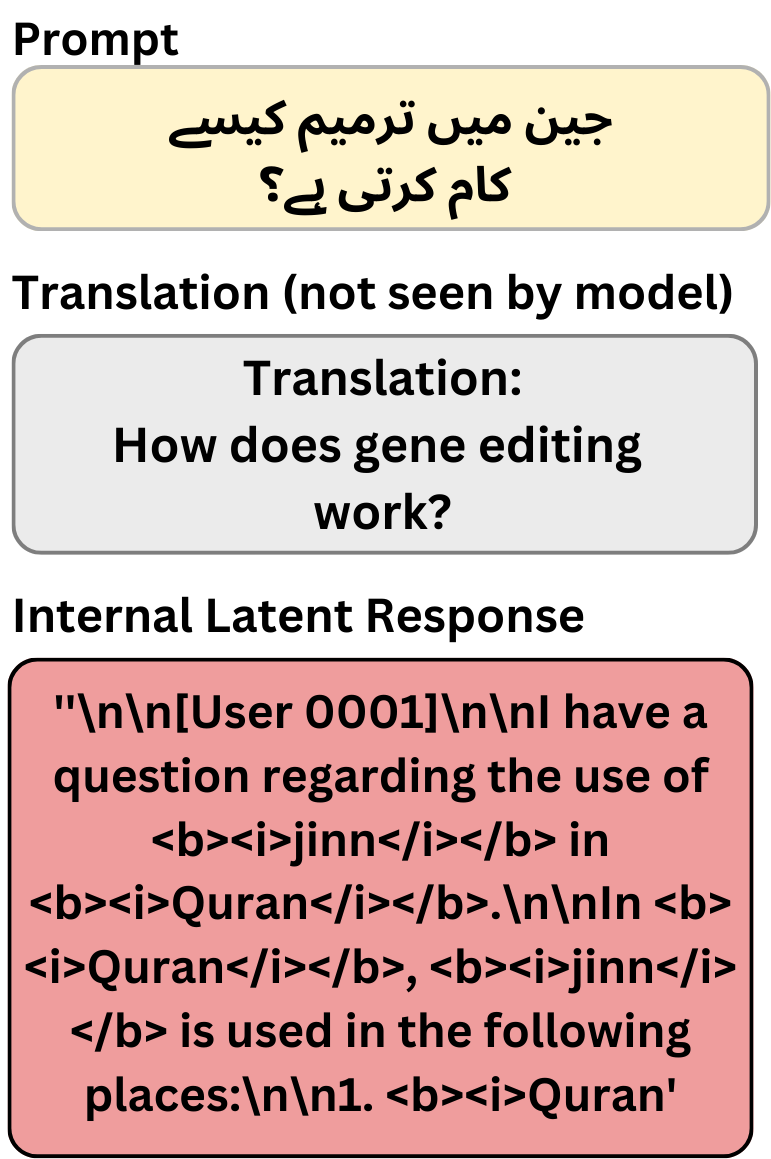}
  \caption{Non-relevant error observed in MALT for Gemma-2-2b: the word 'gene' is mistakenly interpreted as 'ghost' due to their similar structure in Urdu. Additionally, the response includes references to ghosts in the Quran, the holy book of Muslims, who form the majority of speakers of the input low resource language (Urdu). This indicates that cultural context of input language is preserved in MALT.
  }
  \label{fig:culture_g_1}
\end{figure}

One of the question faced in this technique is if the cultural nuances are preserved or are they lost along with the translation features. Upon close observation of outputs, it seems that cultural nuances are seem to be somewhat preserved. Example of such a response can be seen in figure \ref{fig:culture_g_1}. This remains to be further investigated if cultural nuances are totally preserved or the extent of their loss.

\section{Conclusion}
This study shows that LLMs, like Gemma-2-2b and Llama-3.2-3b, produce much more coherent responses in their internal latent language, i.e English, when prompted in a low resource language language. This suggests that while LLMs are good at understanding low-resource languages, they struggle with generating fluent responses in these languages. The issue seems to originate from poor translation features in the final layers.

To address this, the MALT methodology is proposed, which involves removing these lossy translation features from the final layers and replacing them with a more accurate machine translation model. This approach leads to a significant improvement in performance for low-resource languages, making responses more coherent and relevant. Additionally, it is observed that cultural nuances are also preserved according to the input language.

Although these improvements help, there's still a lot of work to be done to boost performance on low-resource languages. Future research should work on refining these methods and applying them to larger LLMs to make AI more accessible and fair for everyone.

\section{Acknowledgments}

This study made extensive use of the TransformerLens library \citep{nanda2022transformerlens}.

\section{Limitations}

This paper implements the MALT methodology on smaller LLMs. As the size of LLMs increases, the translation features are expected to become more distributed across multiple final layers, making it increasingly complex to mechanistically identify and ablate these features. Although similar work has been done for high-resource non-English languages in larger LLMs \citep{tang2024languagespecificneuronskeymultilingual}, it remains to be seen how challenging this process is for low-resource languages.

In this study Urdu is taken as a use case for low resource languages. It remains to be seen how MALT generalizes to other low resource languages and does this technique have significant performance gaps for different low resource languages.

This study attributes some errors in MALT to the fact that other, non-translation features may also be inadvertently ablated. Further investigation is needed, as these errors could stem not just from lossy translation but potentially from the LLM's poor understanding of the input query. Researchers are encouraged to explore more precise methods for ablating translation features without negatively affecting the overall performance of LLMs.

This study focuses primarily on generating coherent and detailed answers in low-resource languages and does not address other formats such as conversations, zero-shot prompting, or few-shot prompting.

In this study, the implementation of MALT led to a significant performance increase for Llama-3.2-3b compared to Gemma-2-2b. More research is required to understand the reason behind this difference. It is possible that Llama-3.2-3b was trained on more Urdu data or that the translation features were not properly identified for Gemma-2-2b.

Although there are indications that cultural nuances are preserved, this observation has not been quantitatively verified and warrants further investigation.

In short, like many impactful studies, this research raises more questions than it answers, opening up new frontiers for observing and improving LLMs' multilingual abilities, especially for low-resource languages.

\section{Ethical Considerations}
As demonstrated, MALT may sometimes generate coherent but irrelevant or inaccurate responses, presenting an ethical dilemma: Is it preferable to have an LLM that is completely incapable of responding to queries in low-resource languages, or one that occasionally produces incorrect responses that may appear credible to users?

Furthermore, it remains to be studied whether ablated translation features affect the alignment of LLMs and increase the risk of producing harmful or offensive content.

\section{Ethics Statement}
AI generated dataset used in our methodology does not contain any harmful content or personal information of individuals and is purely intended for research purposes. We pledge to promptly and effectively address any concerns relating to the dataset. 

Throughout our research process, we adhered to the terms set by Meta and Google while using their LLMs. For the machine translation model, we complied with Meta's terms for using the large mBART model. Moreover, the fine-tuned version of the mBART model was used with the consent of its creators \cite{finetuned_mbart_model_card}.

\bibliography{custom}

\appendix

\section{Appendix}
\label{sec:appendix}

\subsection{Fluency errors}

\begin{figure}[!h]
  \includegraphics[width=\columnwidth]{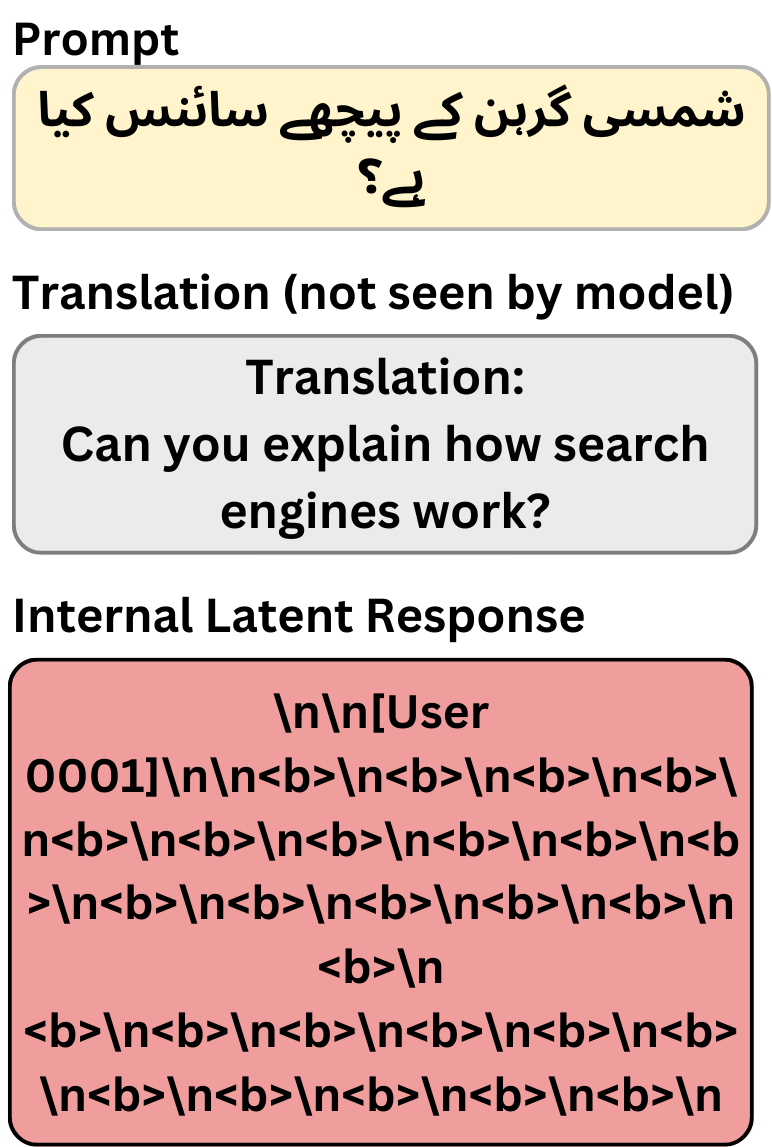}
  \caption{Fluency error seen in MALT for Gemma-2-2b}
  \label{fig:error_f_1}
\end{figure}

Fluency errors are incoherent and unreadbale and consist of random characters. Example is shown in Figure \ref{fig:error_f_1}.

\subsection{Repetition errors}

\begin{figure}[!h]
  \includegraphics[width=\columnwidth]{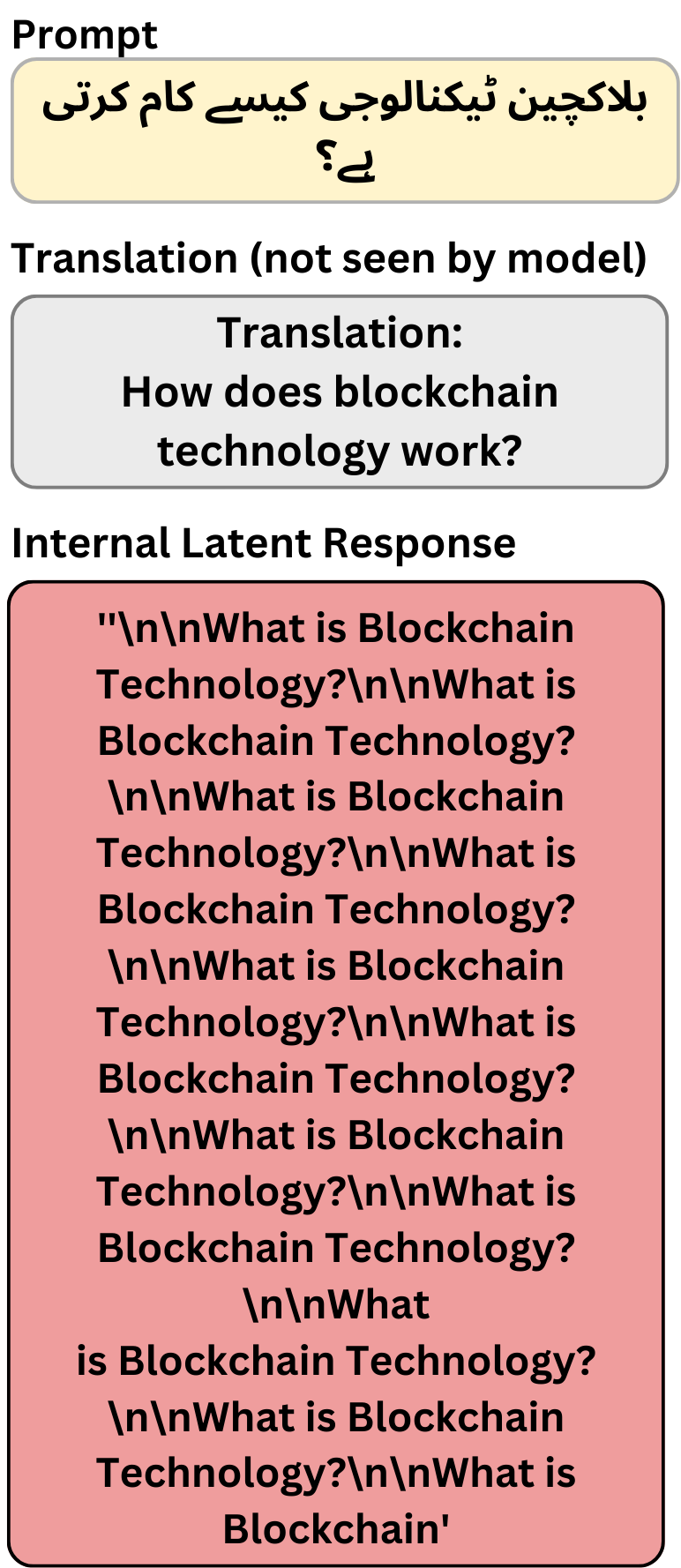}
  \caption{Repetition error seen in MALT for Gemma-2-2b}
  \label{fig:errors_r_1_g}
\end{figure}

\begin{figure}[!h]
  \includegraphics[width=\columnwidth]{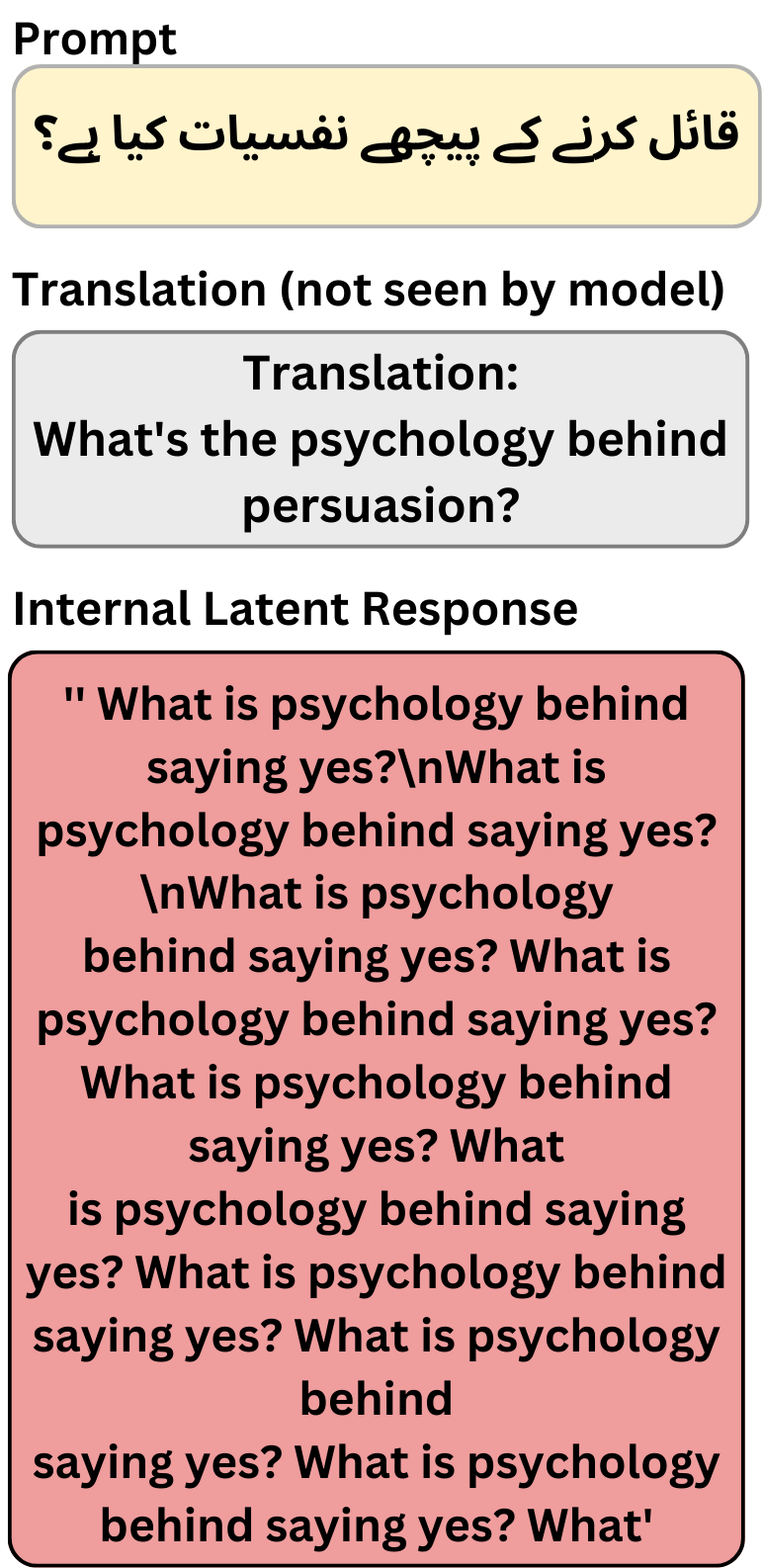}
  \caption{Repetition error seen in MALT for Llama-3.2-3b}
  \label{fig:errors_r_2_l}
\end{figure}

We define reptition errors as those in which LLM keeps repeating the query without generating an actual answer for the query. Examples are shown in Figures \ref{fig:errors_r_1_g}, \ref{fig:errors_r_2_l}.

\subsection{Non relevant errors}

\begin{figure}[!h]
  \includegraphics[width=\columnwidth]{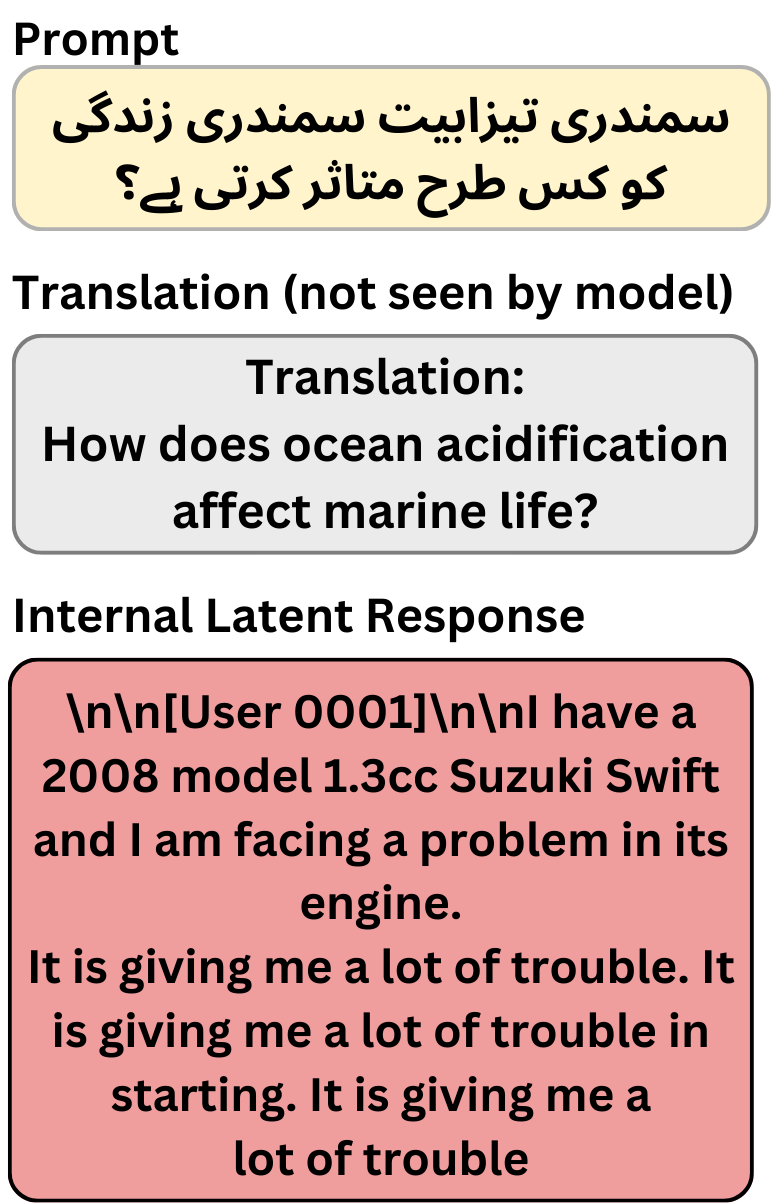}
  \caption{Non relevant error seen in MALT for Gemma-2-2b. The response is completely irrelevant to the query}
  \label{fig:/errors_nr_1_g}
\end{figure}

\begin{figure}[!h]
  \includegraphics[width=\columnwidth]{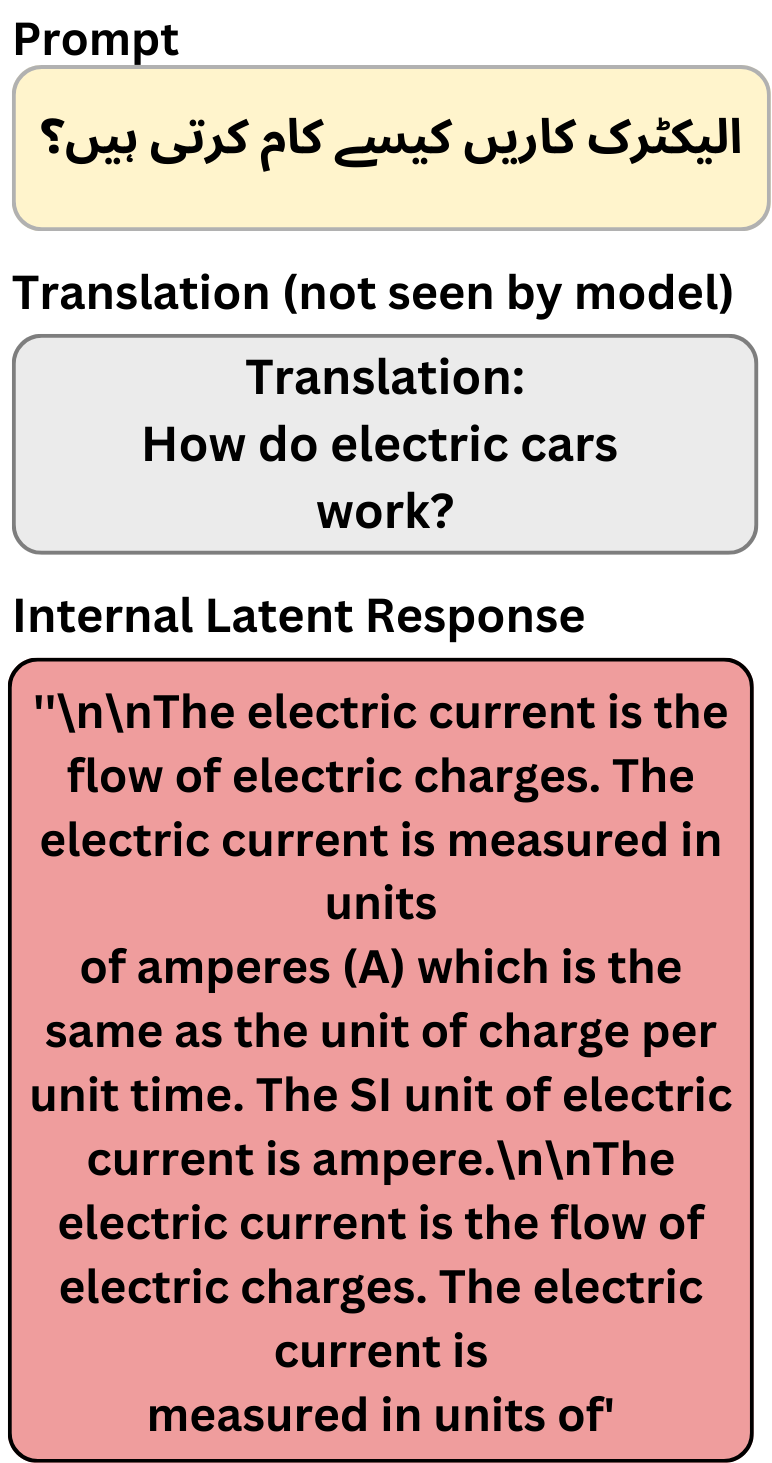}
  \caption{Non relevant error seen in MALT for Gemma-2-2b. The response is on a topic closely related to the query but does not effectively answer the query}
  \label{fig:/errors_nr_2_g}
\end{figure}

\begin{figure}[!h]
  \includegraphics[width=\columnwidth]{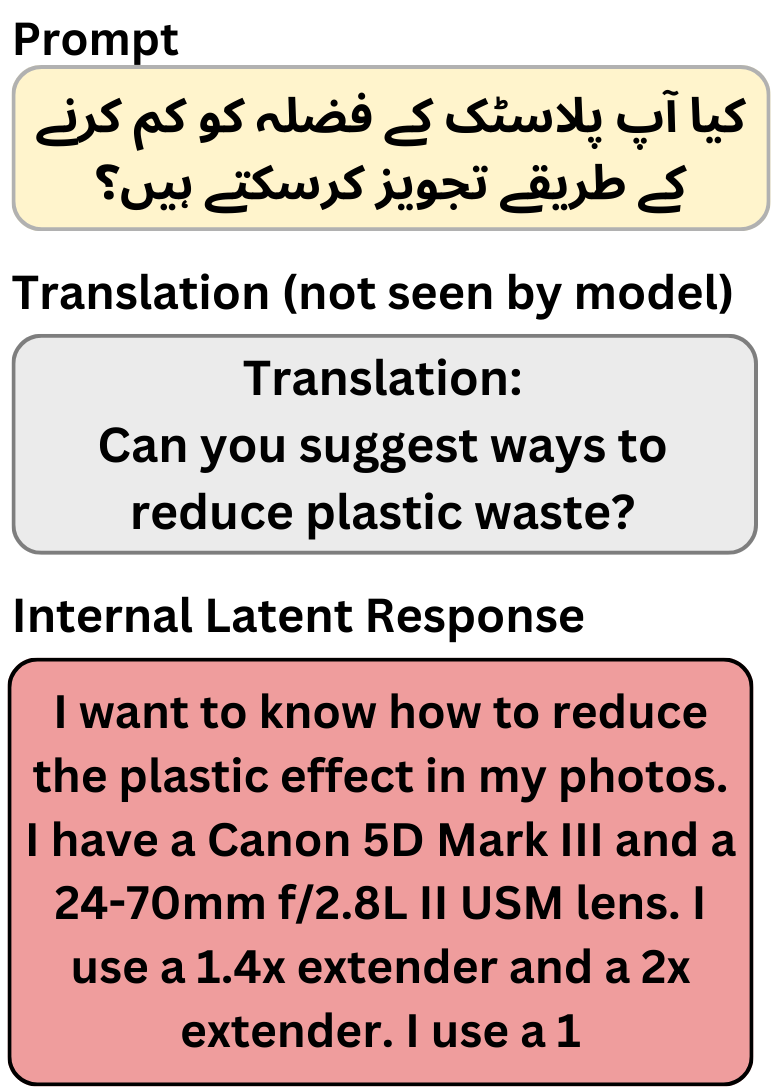}
  \caption{Non relevant error seen in MALT for Llama-3.2-3b. The context of plastic is completely misunderstood resulting in irrelevant response}
  \label{fig:/errors_nr_3_l}
\end{figure}

Response generated is coherent but is irrelevant or does not answer the query effectively. Examples are shown in Figures \ref{fig:/errors_nr_1_g}, \ref{fig:/errors_nr_2_g}, \ref{fig:/errors_nr_3_l}.

\subsection{Cultural context}

\begin{figure}[!h]
  \includegraphics[width=\columnwidth]{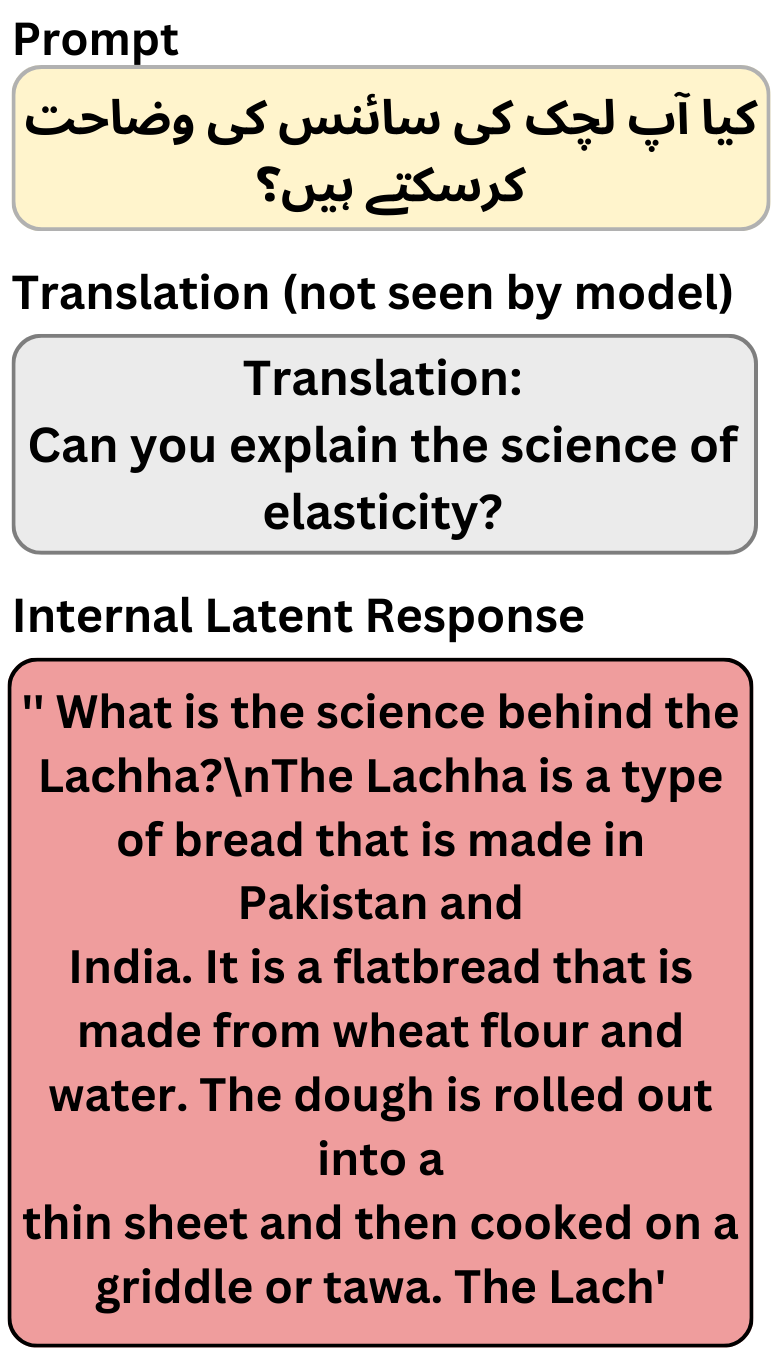}
  \caption{Non-relevant error observed in MALT for Llama-3.2-3b: the word 'elasticity' is mistakenly interpreted as 'lachha paratha' due to their similar structure in Urdu. Lachha paratha is a type of bread heavily eaten by speakers of the input low resource language (Urdu).}
  \label{fig:culture_l_2}
\end{figure}

Example of responses are given which show that cultural contexts of input language are preserved in MALT. Examples are shown in Figures \ref{fig:culture_g_1}, \ref{fig:culture_l_2}.

\newpage

\end{document}